\pdfoutput=1
\documentclass[sigconf,screen]{acmart}

\usepackage[normalem]{ulem}

\newcommand{\argmax}{\mathop{\mathrm{argmax\,}}}

\usepackage{bbm}
\usepackage[ruled, algo2e, vlined]{algorithm2e}

\usepackage{tcolorbox}

\SetCommentSty{mycommfont}

\theoremstyle{definition}

\usepackage{multirow}
\usepackage{tabularx}

\usepackage{pifont}

\allowdisplaybreaks

\AtBeginDocument{%
  \providecommand\BibTeX{{%
    \normalfont B\kern-0.5em{\scshape i\kern-0.25em b}\kern-0.8em\TeX}}}

\setcopyright{acmcopyright}
\copyrightyear{2023}
\acmYear{2023}
\setcopyright{acmlicensed}\acmConference[WWW '23]{Proceedings of the ACM Web Conference 2023}{May 1--5, 2023}{Austin, TX, USA}
\acmBooktitle{Proceedings of the ACM Web Conference 2023 (WWW '23), May 1--5, 2023, Austin, TX, USA}
\acmPrice{15.00}
\acmDOI{10.1145/3543507.3583346}
\acmISBN{978-1-4503-9416-1/23/04}

\begin{document}

\title{Active Learning from the Web}

\author{Ryoma Sato}
\email{r.sato@ml.ist.i.kyoto-u.ac.jp}
\affiliation{%
  \institution{Kyoto University / RIKEN AIP}
}

\renewcommand{\shortauthors}{Ryoma Sato}

\begin{abstract}
Labeling data is one of the most costly processes in machine learning pipelines. Active learning is a standard approach to alleviating this problem. Pool-based active learning first builds a pool of unlabelled data and iteratively selects data to be labeled so that the total number of required labels is minimized, keeping the model performance high. Many effective criteria for choosing data from the pool have been proposed in the literature. However, how to build the pool is less explored. Specifically, most of the methods assume that a task-specific pool is given for free. In this paper, we advocate that such a task-specific pool is not always available and propose the use of a myriad of unlabelled data on the Web for the pool for which active learning is applied. As the pool is extremely large, it is likely that relevant data exist in the pool for many tasks, and we do not need to explicitly design and build the pool for each task. The challenge is that we cannot compute the acquisition scores of all data exhaustively due to the size of the pool. We propose an efficient method, Seafaring, to retrieve informative data in terms of active learning from the Web using a user-side information retrieval algorithm. In the experiments, we use the online Flickr environment as the pool for active learning. This pool contains more than ten billion images and is several orders of magnitude larger than the existing pools in the literature for active learning. We confirm that our method performs better than existing approaches of using a small unlabelled pool.
\end{abstract}



\maketitle

\section{Introduction}

The revolution of deep learning has greatly extended the scope of machine learning and enabled many real-world applications such as computer-aided diagnosis \cite{chen2016mitosis, du2018breast,yu2018recurrent}, anomaly detection, \cite{akcay2018ganomaly,ruff2020deep} and information retrieval \cite{mcauley2015image,song2016deep}. However, the high cost of annotating data still hinders more applications of machine learning. Worse, as the number of parameters of deep models is larger than traditional models, it consumes more labels and, thus, more costs \cite{hestness2017deep,kaplan2020scaling,rosenfeld2020constructive,zhai2022scaling,geirhos2021partial}. Many solutions have been proposed for alleviating the cost of labeling.

First, many unsupervised and self-supervised methods that do not require manual labeling have been proposed \cite{chen2020simple,he2020momentum,baevski2020wav2vec}. However, the applications of pure unsupervised methods are still limited, and many other applications still call for human annotations to some extent. Specifically, unsupervised representation learning offers task-agnostic representations, and we need to extract task-specific information from them using labeled data.

Another popular approach is pre-training. Large scale supervised datasets such as ImageNet \cite{russakovsky2015imagenet} and JFT-3B \cite{zhai2022scaling} and/or unsupervised and self-supervised methods \cite{chen2020simple,devlin2019bert} can be used for pre-training. These methods significantly reduce the number of required labels. However, we still need several labels, even with such techniques. These techniques are orthogonal to the problem setting considered in this paper, and our approach can be used with such pre-training techniques. In particular, we focus on labeling data after applying such techniques that reduce the sample complexity.

The most direct approach to reducing the burden of labeling would be active learning \cite{settles2009active,lewis1994sequential,tong2001support,gal2017deep}. Among many approaches to active learning \cite{angluin1987queries,atlas1989training}, we focus on pool-based active learning, which gathers unsupervised data and iteratively selects data to be labeled so that the number of required labels is minimized. Many pool-based active learning methods have been proposed for decades \cite{lewis1994sequential,tong2001support,gal2017deep,raj2022convergence}. Most researches on active learning focus on the procedure after collecting the pool of data; in particular, most works focus on the criterion of selecting data, i.e., the acquisition function of active learning \cite{gal2017deep,joshi2009multi,hoi2006batch}. However, how to define the pool of data has been less explored. Collecting unsupervised data is not straightforward. Naive crawling on the Web may collect only negative or irrelevant samples and harms the performance of active learning greatly. Making a very large pool on a local server may consume a lot of communication costs and time, and it is not economical or infeasible to build a very large pool in the traditional approach. For example, we build a middle-size pool with $10^5$ images in the experiments, and it consumes about 20 GB of storage. Building a ten or hundred times larger pool on a local server or on cloud storage may not be economical, especially when the target task is small. The cost scales linearly with the size of the pool. It is infeasible to host a pool with $10^{10}$ images in any case.

\begin{figure*}[tb]
\centering
\includegraphics[width=0.85\hsize]{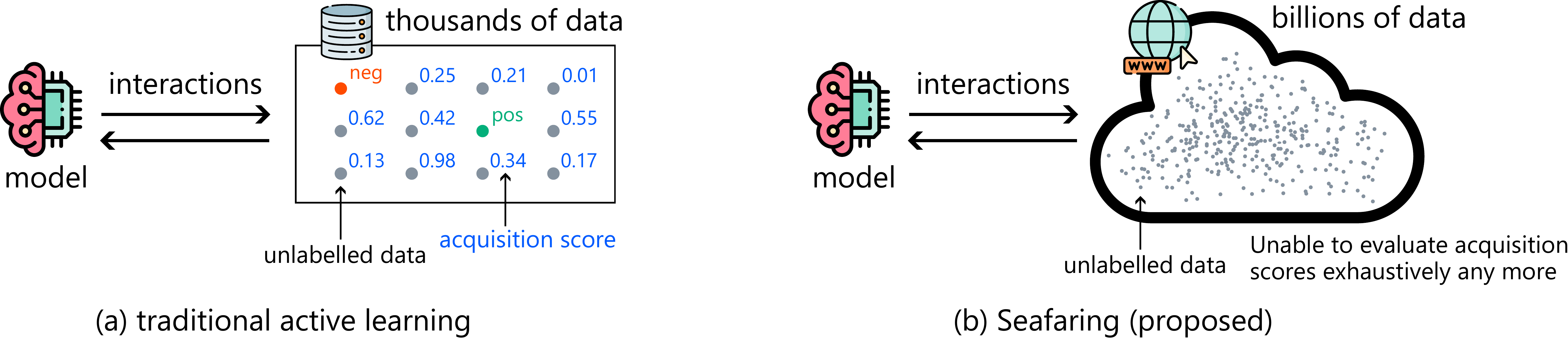}
\vspace{-0.1in}
\caption{\textbf{Illustrations of traditional active learning and Seafaring.} Traditional methods use a local database as the pool of active learning. Seafaring uses a huge database on the Internet as the pool of active learning. As the pool of Seafaring is extremely large, it is more likely that informative data exist, but it is challenging to find it efficiently.}
\label{fig: illustration}
\vspace{-0.1in}
\end{figure*}

\begin{table}[tb]
    \centering
    \caption{Notations.}
    \vspace{-0.1in}
    \begin{tabular}{ll} \toprule
        Notations & Descriptions \\ \midrule
        $\mathcal{X}$ & The input space, i.e., images. \\
        $\mathcal{Y}$ & The label space, i.e., $\{0, 1\}$. \\
        $\mathcal{U} \subset \mathcal{X}$ & The pool of active learning. \\
        $\mathcal{U}_{\text{Flickr}} \subset \mathcal{X}$ & The set of all images on Flickr. \\
        $\mathcal{L} \subset \mathcal{X}$ & The set of initial labelled data. \\
        $\mathcal{O}\colon \mathcal{X} \to \mathcal{Y}$ & The labeling oracle. \\
        $f_{\theta}\colon \mathcal{X} \to [0, 1]^\mathcal{Y}$ & The classification model. \\
        $a\colon \mathcal{X} \to \mathbb{R}$ & The acquisition function. \\
        $B \in \mathbb{Z}_+$ & The labeling budget of active learning. \\
        \bottomrule
    \end{tabular}
    \label{tab: notations}
\end{table}

In this paper, we regard items on the Web as a very large pool of unlabelled data and advocate the use of this large pool for active learning (Figure \ref{fig: illustration}). In the experiments, we use all the images on Flickr as the pool. This pool contains more than ten billion ($10^{10}$) images, which is several orders of magnitude larger than the pools that have ever been seen in the literature. As the pool is extremely large, it is likely that informative items exist in the pool. Besides, our approach does not require the user to explicitly build the pool on a local server before the application of active learning, but it retrieves informative items on the fly during active learning. Therefore, it further reduces the cost of preparation of data in addition to the reduction of the number of required labels.

However, they are many challenges to realizing this goal.
\begin{itemize}
\item As there are too many items in the pool, it is infeasible to compute the scores of all items exhaustively.
\item We do not even have the list of items on the Web or access to the database directly. The only way we access items is via the use of limited search queries.
\end{itemize}
We propose Seafaring (\underbar{Sea}rching \underbar{f}or \underbar{a}ctive lea\underbar{r}n\underbar{ing}), which uses a user-side information retrieval algorithm to overcome these problems. User-side search algorithms were originally proposed for building customized search engines on the user's side \cite{bra1994information,sato2022retrieving}. In a nutshell, our proposed method uses the acquisition function of active learning as the score function of the user-side engine and retrieves informative items to be labeled efficiently without direct access to external databases.

The contributions of this paper are summarized as follows.
\begin{itemize}
\item We advocate the use of items on the Web as the pool of active learning. This realizes several orders of magnitude larger pools than existing approaches and enables us to find informative examples to be labeled.
\item We propose Seafaring, an active learning method that retrieves informative items from the Web.
\item We show the effectiveness of our proposed methods with a synthetic environment and real-world environment on Flickr, comparing with the existing approach that uses a small pool.
\item We show a supervising effect of our simple approach, i.e., Seafaring retrieves positive labels from a myriad of irrelevant data.
\end{itemize}

\begin{tcolorbox}[colframe=gray!20,colback=gray!20,sharp corners]
\textbf{Reproducibility}: Our code is publicly available at \url{https://github.com/joisino/seafaring}.
\end{tcolorbox}

\section{Problem Setting} \label{sec: setting}

\begin{table}[tb]
    \centering
    \caption{Size of the pools of unlabelled data. The pools of existing works contain hundreds of thousands of data. The pool of Seafaring is several orders of magnitude larger than pools that have ever been seen in the literature.}
    \vspace{-0.1in}
    \begin{tabular}{ll} \toprule
        Dataset & Pool Size \\ \midrule
        Reuters \cite{tong2001support} & 1,000 \\
        NewsGroup \cite{tong2001support} & 500 \\
        ImageCLEF \cite{hoi2006batch} & 2,700 \\
        Pendigits \cite{joshi2009multi} & 5,000 \\
        USPS \cite{joshi2009multi} & 7,000 \\
        Letter \cite{joshi2009multi} & 7,000 \\
        Caltech-101 \cite{joshi2009multi} & 1,500 \\
        CACD \cite{wang2017cost} & 40,000 \\
        Caltech-256 \cite{wang2017cost} & 24,000 \\
        MNIST \cite{gal2017deep} \cite{kirsch2019batchbald} & 50,000 \\
        ISBI 2016 \cite{gal2017deep} & 600 \\
        EMNIST \cite{kirsch2019batchbald} & 94,000 \\
        CINIC-10 \cite{kirsch2019batchbald} & 160,000 \\
        MNIST \cite{beluch2018power} & 2,000 \\
        CIFAR-10 \cite{beluch2018power} & 4,000 \\
        CIFAR-10 \cite{beluch2018power} & 20,000 \\
        Diabetic R. \cite{beluch2018power} & 30,000 \\
        ImageNet \cite{beluch2018power} & 400,000 \\
        Synthetic (ours) & 100,000 \\
        Flickr (ours) & $\ge$ 10,000,000,000 \\
        \bottomrule
    \end{tabular}
    \label{tab: size}
\end{table}

\begin{figure*}[tb]
\centering
\includegraphics[width=0.9\hsize]{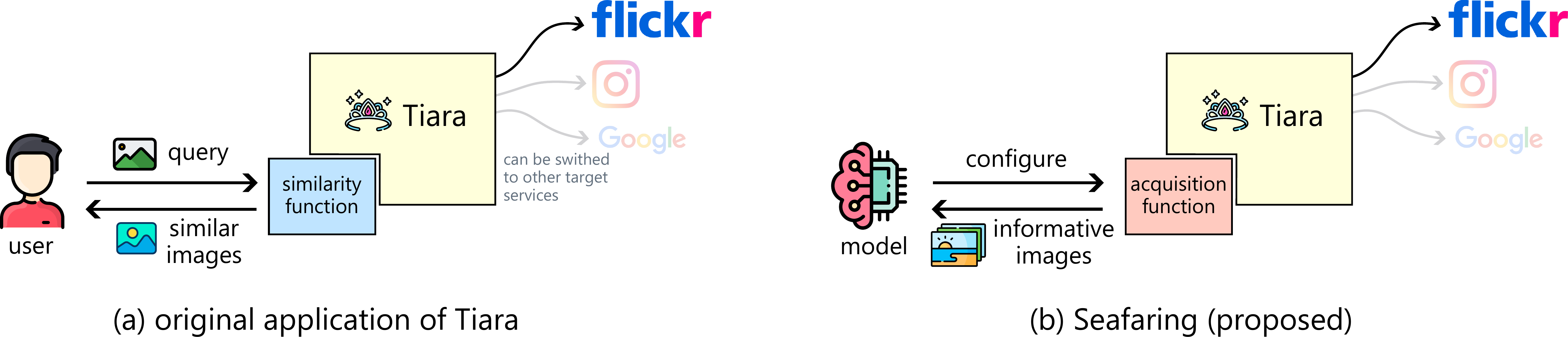}
\vspace{-0.1in}
\caption{\textbf{Illustrations of Tiara and Seafaring.} Originally, user-side search engines and Tiara were proposed to build similar image search engines on the user's side for the Web services without official similar image search engines. The advantages of these methods are that they work with user-defined score functions and they work without any privileged access to the database. Seafaring uses Tiara in active learning settings. Seafaring uses an acquisition function of active learning as the score function of the user-side search engine and retrieves informative images for the current model.}
\label{fig: tiara}
\vspace{-0.1in}
\end{figure*}

Let $\mathcal{X}$ and $\mathcal{Y}$ be the input and label domains, respectively. In this paper, we focus on the binary classification of image data, though our approach can be extended to other tasks. Therefore, $\mathcal{X}$ is the set of images and $\mathcal{Y} = \{0, 1\}$. Let $\mathcal{U} = \{x_i\} \subset \mathcal{X}$ denote the set of unlabelled data, let $\mathcal{L} = \{(x_i, y_i)\} \subset \mathcal{X} \times \mathcal{Y}$ denote the set of initial labelled data. Let $\mathcal{O}\colon \mathcal{X} \to \mathcal{Y}$ be the labeling oracle function, which takes an image as input and outputs the ground truth label. In reality, the oracle $\mathcal{O}$ asks human annotators to reveal the ground truth label. The evaluation of the oracle $\mathcal{O}$ is costly because it requires human resources. We want to evaluate the oracle function as few times as possible. We limit the number of the evaluation by $B \in \mathbb{Z}_+$. $B$ is called the budget of active learning.

Active learning involves $B$ iterations. Let $\mathcal{L}_0 = \emptyset$ be the initial set of selected data. In $i$-th iteration, the model $f_{\theta}$ is trained with the current set of labelled images $\mathcal{L} \cup \mathcal{L}_{i-1}$, and an active learning method selects a unlabelled sample $x_i \in \mathcal{U}$ from the pool. Then, the oracle (e.g., human annotators) reveals the label $y_i = \mathcal{O}(x_i)$. The set of labelled data is updated by setting $\mathcal{L}_i = \mathcal{L}_{i-1} \cup \{(x_i, y_i)\}$. After $B$ iterations, the performance of the model is evaluated with the obtained training data $\mathcal{L} \cup \mathcal{L}_{B}$. The aim of active learning is to obtain a good model with a small $B$.

Typical active learning methods select data based on acquisition function $a\colon \mathcal{X} \to \mathbb{R}$ and set $x_i = \argmax_{x \in \mathcal{U}} a(x)$. The design of the acquisition function depends on the method, and many acquisition functions have been proposed, such as the entropy function \cite{settles2009active}, least confidence \cite{settles2009active}, margin sampling \cite{settles2009active}, expected error reduction \cite{roy2001toward}, expected model change \cite{settles2007multuple}, and ensemble-based scores \cite{beluch2018power}. The family of acquisition functions, including entropy, least confidence, and margin sampling, is sometimes referred to as uncertainty sampling, and they are all equivalent in the binary classification setting \cite{settles2009active,nguyen2022how}. In this paper, we do not pursue the design of the acquisition function but use the existing one. Instead, we focus on the design of the pool $\mathcal{U}$.

\section{Proposed Method}

In this section, we introduce our proposed method, Seafaring, which explores the sea of data on the Internet in search of informative data.

Seafaring uses all the images in an image database on the Internet as the pool $\mathcal{U}$. In the experiments, we use the image database $\mathcal{U}_{\text{Flickr}}$ of Flickr as the pool. It should be noted that our method is general, and other databases such as Instagram $\mathcal{U}_{\text{Instagram}}$ and DeviantArt $\mathcal{U}_{\text{DeviantArt}}$ can be used as the target database depending on the target task. We focus on Flick because it contains sufficiently many images for general tasks, and its API is easy to use in practice. As $\mathcal{U}_{\text{Flickr}}$ contains more than $10$ billion images of various motifs and styles, it is likely that relevant and informative images exist in them. Existing methods use smaller pools, which contain thousands of items (Table \ref{tab: size}). As these pools are task-specific, existing active learning methods need to design the pool for each task. Existing methods for active learning assume that a task-specific pool is available for free, but we emphasize that this is not the case in many cases, and designing and building task-specific pools is costly. For example, suppose we build an image recommender system and train a model that classifies images that a user likes for each user. It is not obvious what kind of images each user likes beforehand, and preparing an appropriate pool of unlabelled data is a difficult task. The pool becomes very large if we gather a pile of images from the Web without any care, and traditional active learning is infeasible in such a case. By contrast, Seafaring handles an extremely large pool, and we can use it for many general tasks without any additional effort. Thus Seafaring further reduces the burden of data preparation. In addition, if we update the model regularly with continual or life-long learning methods \cite{li2016learning,zenke2017continual,wu2019large}, existing active learning methods need to maintain the pool. By contrast, Seafaring does not require the users to maintain the pool because the users of Flickr add new images to the Flickr database $\mathcal{U}_{\text{Flickr}}$ every day, and the pool is maintained up-to-date automatically and autonomously. Thus Seafaring always uses an up-to-date database without any special effort.

In each iteration of active learning, Seafaring searches the best image with respect to the acquisition function $a$ from $\mathcal{U}_{\text{Flickr}}$ and sets $x_i \leftarrow \argmax_{x \in \mathcal{U}_{\text{Flickr}}} a(x)$. However, as $\mathcal{U}_{\text{Flickr}}$ is extremely large, we cannot evaluate $a(x)$ for each image $x \in \mathcal{U}_{\text{Flickr}}$. It is impossible to even download all images in $\mathcal{U}_{\text{Flickr}}$ or even list or enumerate $\mathcal{U}_{\text{Flickr}}$. Employees of Flickr could build bespoke search indices and databases for active learning purposes, but as we and most readers and users of Seafaring are not employees of Flickr, the approach of building an auxiliary index of the Flickr database is infeasible.

To overcome this issue, Seafaring uses a user-side search algorithm, namely, Tiara \cite{sato2022retrieving}. Tiara retrieves items from external databases on the Internet. The original motivation of Tiara is that most users of Web services are not satisfied with the official search engine in terms of its score function or its interface, and Tiara offers a way for users to build their own search engines for existing Web services. Seafaring uses Tiara differently. Seafaring sets the acquisition function $a$ of active learning as the score function of Tiara and thereby retrieves informative images for model training (not for users). Figure \ref{fig: tiara} shows an illustration of Seafaring and the difference between the original application of Tiara and Seafaring. As Tiara works without privileged access to external databases (i.e., $\mathcal{U}_{\text{Flickr}}$), and ordinary users can run it, it suits our situation well.

We introduce the algorithm of Tiara briefly for the completeness of this paper. Tiara takes a score function $s\colon \mathcal{X} \to \mathbb{R}$ as input. The aim of Tiara is to retrieve an image $x^*$ that maximizes $s$ from an external database. Tiara assumes that a tag or text-based search system that takes text as input and returns a set of images is available for the external database. In Flickr, Tiara and Seafaring use \texttt{flickr.photos.search} API\footnote{\url{https://www.flickr.com/services/api/flickr.photos.search.html}} for the tag-based search system. Tiara formulates the search problem as a multi-armed bandit problem where a tag is an arm, and the score function is the reward. Tiara uses LinUCB algorithm \cite{li2010contextual} to solve the multi-armed bandied problem with pre-trained tag features based on GloVe \cite{pennington2014glove}. Intuitively, Tiara iteratively queries tags (exploration), evaluates images, finds promising tags, and finds out (sub-)optimal images by querying promising tags (exploitation). Please refer to the original paper \cite{sato2022retrieving} for more details.

\begin{figure*}[tb]
\centering
\includegraphics[width=0.85\hsize]{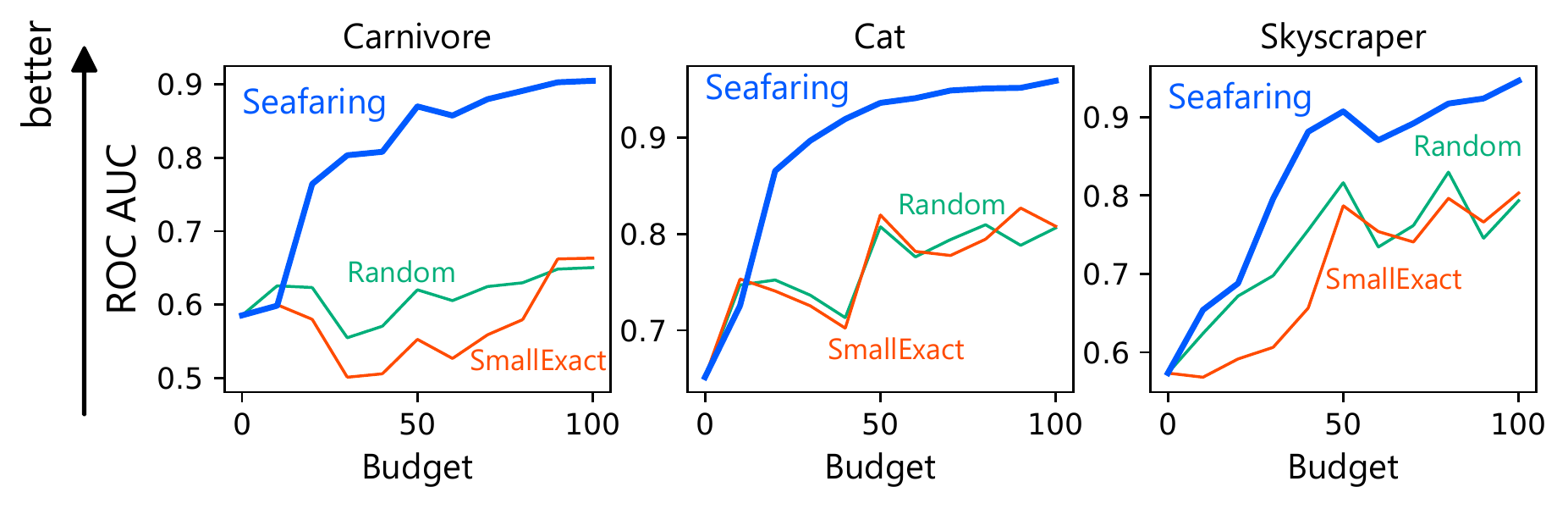}
\vspace{-0.2in}
\caption{\textbf{Results of OpenImage dataset.} Each curve reports the performance of each method. Point $(x, y)$ means that the method achieves $y$ AUC within $x$ access to the label oracle. Higher is better. These results show that Seafaring is more effective than the baseline methods.}
\label{fig: openimage}
\vspace{-0.1in}
\end{figure*}

Seafaring uses the exponential of the entropy function \begin{align} \label{eq: acquisition}
    a(x) = s(x) = \exp\left(- \gamma \sum_{i \in \mathcal{Y}} f_\theta(x)_i \log f_\theta(x)_i\right)
\end{align}
as the acquisition function and the score function of Tiara. $f_\theta(x)_i \in [0, 1]$ is the output probability of class $i$ by the model $f_{\theta}$. $\gamma \in \mathbb{R}$ is a hyperparameter representing the temperature. We confirmed that the algorithm was insensitive to the choice of this parameter and set $\gamma = 4$ throughout the experiments. The entropy function is one of the most popular acquisition functions of active learning \cite{holub2008entropy,nguyen2022how}. Intuitively, if the entropy function is high, the model $f_\theta$ is not certain about the label of this data, so labeling this data is likely to provide much information for the model. Our acquisition function $a$ is monotone with respect to the entropy function. We apply the exponential function because data with high scores are more important than data with low scores, e.g., whether the entropy is $0.8$ or $0.9$ is more important than whether the entropy is $0.2$ or $0.3$ for searching high entropy data. The exponential function magnifies the difference in the high score regime. It should be noted that monotone transformation does not change the results if we evaluate the scores of all data and select the exact maximum data, but the results of Seafaring change by a monotone transformation because of the exploration-exploitation trade-off of the bandit algorithm and the inexact search scheme. Seafaring is general and can be combined with any other acquisition functions such as the ensemble score \cite{beluch2018power} and BALD \cite{gal2017deep}. As the design of the acquisition function is not the primal interest of this paper, we leave further exploration of acquisition functions for future work.

Seafaring is indeed a simple application of Tiara to active learning. We emphasize that the formulation of Web-scale active learning is the core interest of this paper, and the optimization method is secondary. Note also that we tried a more complicated method in preliminary experiments, but we found out that the vanilla method was already effective and had preferable features, as we analyze in the experimental section. So we keep our proposed method as simple as possible for usability and customizability.

\section{Experiments}

We show that Seafaring is more effective than the traditional approach that uses a small and fixed pool of unlabelled data. We conduct control experiments where the baseline method uses the same settings to contrast a huge pool with a small pool. Specifically, we let the baseline method use the same acquisition function and samples from the same database as those of Seafaring. In the analysis, we also show that Seafaring automatically balances the ratio of labels without any special mechanism.

\subsection{Experimental Setups}

We aim at training ResNet-18 \cite{he2016deep} for various tasks. One shot of an experiment is conducted as follows: Initially, one positive sample and one negative sample are given, and they form the initial training set $\mathcal{L}$. Through the experiments, an active learning method can evaluate the oracle $\mathcal{O}$ at most $B$ times. We set $B = 100$ throughout the experiments. In a loop of an experiment, the ResNet model $f_{\theta}$ is initialized by the ImageNet pre-trained weights\footnote{\texttt{ResNet18\_Weights.IMAGENET1K\_V2} in \url{https://pytorch.org/vision/stable/models.html}.}, and then trained with $\mathcal{L}$. The training procedure uses the stochastic gradient descent (SGD) with learning rate $0.0001$ and momentum $0.9$ and with $100$ epochs. Then, the evaluation program measures the accuracy of the model using test data. The accuracy of the model is measured by ROC-AUC. Note that the results of the evaluation are not fed back to the active learning method, but only the evaluator knows the results so as not to leak the test data. After the evaluation, an active learning method selects one unlabeled sampled $x_i$ from the pool using the trained model and an acquisition function. After the selection of the sample, the label oracle $\mathcal{O}(x_i)$ is evaluated and the label $y_i$ is notified to the active learning method, and $(x_i, y_i)$ is inserted to $\mathcal{L}$. This loop is continued until the method exhausts the budget, i.e., $B$ times. The goal of an active learning method is to maximize the accuracy of the model $f_{\theta}$ during the loop and at the end of the loop. 

It should be noted that the ResNet-18 model is relatively larger than the models used in existing works \cite{wang2017cost,gal2017deep}, where shallow multi-layered perceptron and convolutional neural networks are used, and it further introduces the challenge in addition to the huge pool because each evaluation of the acquisition function costs more.

We use two baseline methods. The \textbf{SmallExact} method is an active learning method with the traditional approach. Specifically, it first samples $1000$ unlabelled data from the database that Seafaring uses, i.e., $\mathcal{U}_{\text{Flickr}}$ or $\mathcal{U}_{\text{OpenImage}}$, where $\mathcal{U}_{\text{OpenImage}}$ is the pool of the synthetic database we will use in the next subsection. SmallExact uses these data for the pool of active learning. It uses the same acquisition function (Eq. \eqref{eq: acquisition}) as Seafaring to make the conditions the same and to clarify the essential difference between our approach and the existing approach. The only differences between Seafaring and SmallExact are (i) the size of the unlabelled pool and (ii) Seafaring carries out an inexact search, while SmallExact evaluates the scores of the data in the pool exhaustively and exactly selects the best data from the pool. The exact search approach of SmallExact is infeasible if the pool or the model is large. In the following, we show that Seafaring performs better than SmallExact even though Seafaring does not carry out an exact search. The \textbf{Random} method selects a random sample from the same pool of Seafaring uses, i.e., $\mathcal{U}_{\text{Flickr}}$ or $\mathcal{U}_{\text{OpenImage}}$, in each loop. Both Seafaring and Random use the same pool, but Random does not use active learning. The difference between the performances of Random and Seafaring shows the benefit of active learning.

We conduct the experiments on NVIDIA DGX-1 servers, and each experiment runs on a V100 GPU.

\subsection{OpenImage Dataset}

\begin{figure*}[tb]
\centering
\includegraphics[width=0.85\hsize]{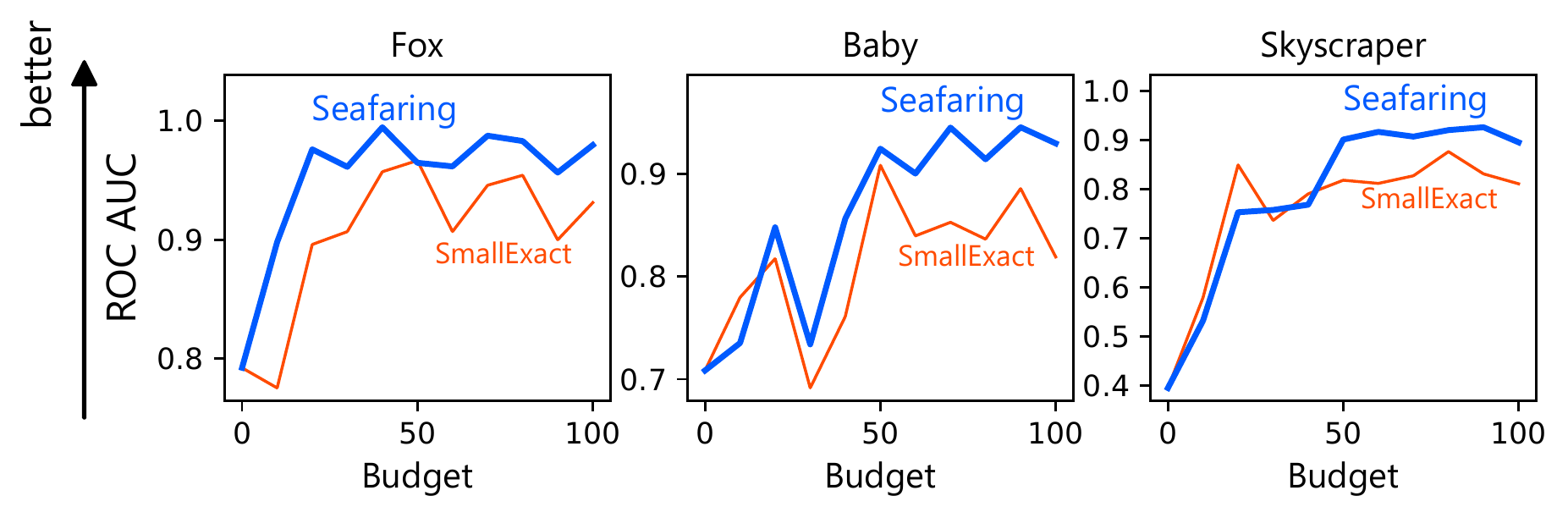}
\vspace{-0.1in}
\caption{\textbf{Results of Flickr dataset.} Each curve reports the performance of each method. Higher is better. These results show that Seafaring is more effective than the traditional approach with a small pool in the real-world Flickr environment.}
\label{fig: flickr}
\vspace{-0.1in}
\end{figure*}

We first construct a synthetic and middle-size image database using the OpenImage Dataset \cite{kuznetsova2020open} to quantitatively evaluate the methods, confirm the general trends, and analyze the results. We also aim to make the experiments easy to reproduce by keeping the size of the environment not so large. We first sample $100\,000$ images from the OpenImage Dataset. Let $\mathcal{U}_{\text{OpenImage}}$ denote the set of these images. It should be noted that even $\mathcal{U}_{\text{OpenImage}}$ is larger than many of the pools used in existing work (Table \ref{tab: size}). We use the tag annotations of the OpenImage Dataset to construct a virtual search engine for Tiara, following the original paper \cite{sato2022retrieving}. We define a classification task with a tag $t$. In each task, the images with the tag $t$ are positive samples, and the other images are negative samples. The goal of the task is to train a model that classifies positive and negative samples accurately.

We set the number of iterations of LinUCB in Tiara to $1000$ in this experiment. The number indicates the number of evaluations of the model $f_{\theta}$ and the number of search queries in each loop of active learning. Although the labeling oracle $\mathcal{O}$ is the most costly operation, the evaluation of the model $f_{\theta}$ is also costly when the model is large. So it is desirable to keep this number small. As the SmallExact method evaluates the acquisition scores of all of the $1000$ samples in the pool in each iteration, the number of evaluations of the model is the same for both methods in this experiment.

We run each task $5$ times with different random seeds and report the average performances. Figure \ref{fig: openimage} shows the accuracies of Seafaring and the baseline methods for three tasks, t = Carnivore, t = Cat, and t = Skyscraper. Seafaring outperforms both baselines in all tasks. In t = Carnivore, both baselines fail to find relevant images at all, and the performance does not increase with more labels. By contrast, Seafaring efficiently finds out informative data, and the performance quickly grows with small labels in all tasks.

\subsection{Flickr Environment}

We use the real-world Flickr database $\mathcal{U}_{\text{Flickr}}$ as the pool in this experiment. As we have mentioned before, $\mathcal{U}_{\text{Flickr}}$ contains more than ten billion images, and this is several orders of magnitude larger than the largest pool of active learning in the literature (Table \ref{tab: size}). We emphasize that we are not employees of Flickr nor have privileged access to the Flickr database. Nevertheless, we show that Seafaring retrieves informative images from the Flickr database.

We define a classification task with a tag $t$ but in a different way from the OpenImage experiment. Specifically, we gather $10$ images  $\mathcal{X}_t$ with tag $t$ from the OpenImage dataset, extract features of images using pre-trained ResNet18, and define images whose maximum cosine similarity between $\mathcal{X}_t$ is larger than a threshold as positive samples, and other images as negative samples. Note that $\mathcal{X}$ is not revealed to us and the learner, but only the evaluator knows it. Intuitively, images with features of tag $t$ are positive samples, and other images are negative samples. It should be noted that Flickr also provides tags for images, but we found that they are sometimes missing and noisy. For example, some skyscraper images are not tagged as a skyscraper, i.e., positive samples are wrongly labeled as negatives, and the evaluation becomes not reliable. So we do not use the tags in Flickr to define the task but adopt the similarity-based tasks. 

In this experiment, we set the number of iterations of LinUCB to $100$ to reduce the number of API queries to the Flicker server. As the SmallExact method cannot reduce the number of evaluations but conducts an exact search, we keep the number of evaluations of the SmallExact method $1000$. Therefore, in this experiment, the SmallExact method evaluates the model $10$ times many times in each loop of active learning, and the setting is slightly advantageous to the SmallExact method. We show that Seafaring nevertheless performs better than SmallExact.

Figure \ref{fig: flickr} shows the accuracies of Seafaring and SmallExact for three tasks, t = Skyscraper, t = Fox, and t = Baby. Seafaring outperforms SmallExact in all tasks.

\begin{figure}[tb]
\centering
\includegraphics[width=0.85\hsize]{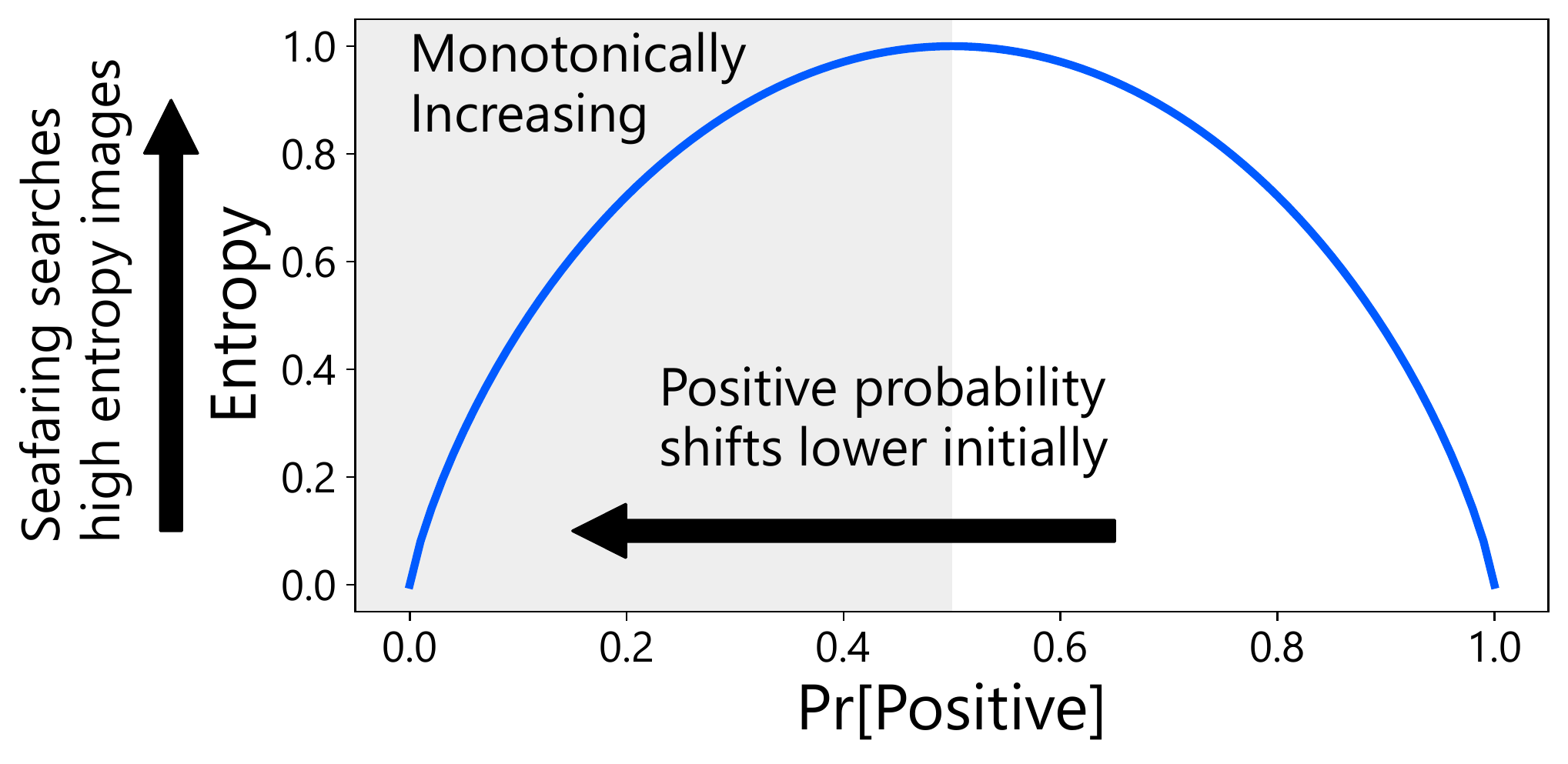}
\vspace{-0.1in}
\caption{\textbf{Illustration of the entropy function.} In the region of $\text{Pr}[\text{Positive}] \le 0.5$, the entropy function is monotonically increasing. In the initial phase of active learning, the model's output probability shifts lower, and the positive probability is less than $0.5$ for most data. Maximizing the entropy is equivalent to maximizing the positive probability in this regime.}
\label{fig: entillust}
\vspace{-0.1in}
\end{figure}

\subsection{Analysis: Seafaring Retrieves Positive Data When They Lack}

As the pool is huge and diverse, most data in the pool are irrelevant to the task, and only a fraction of the data are positive or difficult negative cases. If we randomly sample data from the pool, the obtained labels are negative with high probability, and we obtain little information. For example, in the classification of Carnivore images, most data, such as airplane, car, and building images, are irrelevant to the task. We want to gather relevant images such as cat images (positive samples) and horse images (difficult negative samples). However, it is difficult to distinguish relevant samples from other data until the model's performance is high. Gathering relevant samples and training good models are like a chicken and egg problem. Besides, as the pool contains much more irrelevant images than relevant images, it is all the more difficult to filter out irrelevant images. Let us clarify the difficulty. Suppose we have an auxiliary model that classifies whether a sample is relevant to the task, and this model's accuracy is $0.99$. As there are $10^{10}$ irrelevant images in the pool, it classifies $10^{10} \times (1 - 0.99) = 10^8$ irrelevant data as relevant. If there are $10^4$ relevant images in the pool, the morel classifies $10^4 \times 0.99$ relevant data as relevant. Thus, the majority of the selected data are irrelevant regardless of the high accuracy of the model. It indicates that finding out relevant data is crucial for active learning methods when the pool is huge and unbalanced.

Seafaring does not have an explicit mechanism to cope with this challenge. Nevertheless, we found that Seafaring implicitly solves this problem. Seafaring indeed struggles with finding relevant data in the initial iterations, and the ratio of negative data to positive data grows drastically. After several iterations, most data Seafaring has are negative. The target labels of the training data with which the model $f_{\theta}$ is trained are negative, and the model's output drastically tends to be negative. In other words, the model becomes very confident that most data are negative. Recall that the acquisition function is high when the model's output is not confident, and the entropy function and the acquisition function monotonically increase as the positive probability increases in the regime of the low positive probability (Figure \ref{fig: entillust}). Therefore, in this regime, low confidence and high positive probability are equivalent, and Seafaring searches for data that are likely to be positive. We observe that Seafaring once enters this phase in the initial phase and then finds some positive samples by maximizing the positive probability, the data become balanced, the model becomes accurate, and after that, Seafaring searches informative data with which the accurate model is confused.

\begin{figure*}[tb]
\centering
\includegraphics[width=0.8\hsize]{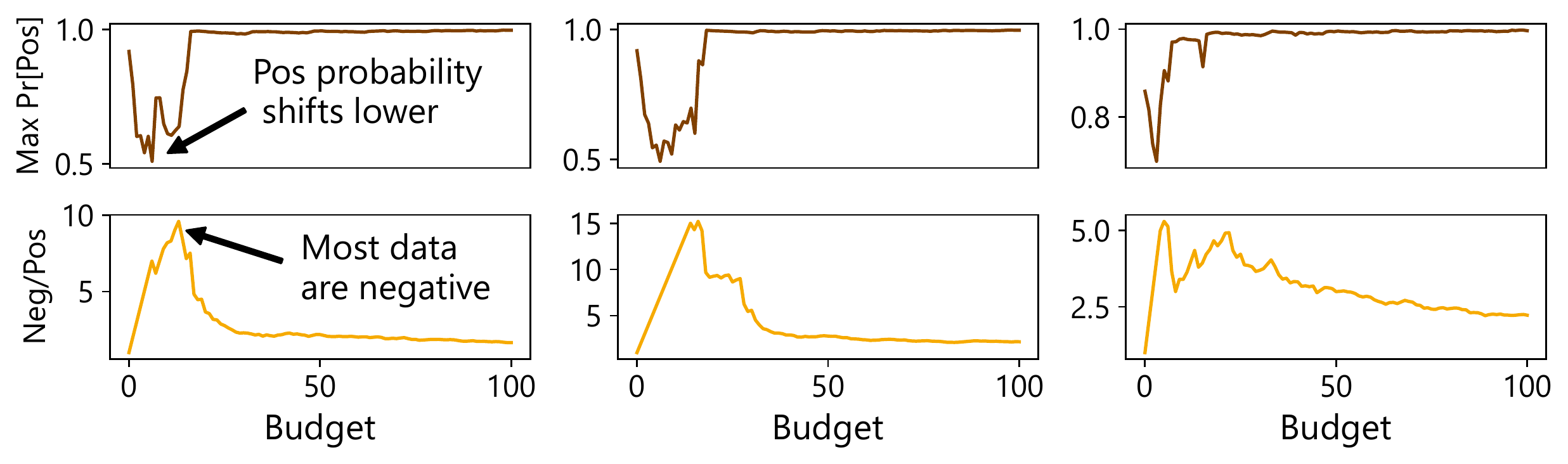}
\vspace{-0.1in}
\caption{\textbf{Transition of maximum positive probability and the ratio of positive and negative samples of Seafaring.} (Top) The maximum positive probability of the data Seafaring evaluates for each loop. In most iterations, the highest probability is around $1.0$, and the curve sticks to the top of the plot. But in the initial phase, even the highest positive probability is around $0.5$ because the model's outputs stick to negative. (Bottom) The ratio of the number of negative samples to the number of positive samples in the labeled data $\mathcal{L}$. Seafaring finds only negative samples at the beginning of iterations, and the ratio grows rapidly. Seafaring searches images with the highest positive probability in this regime. After some positive samples are found, the ratio drops and stabilizes. }
\label{fig: entropy}
\end{figure*}

Figure \ref{fig: entropy} shows the transition of statistics in the course of active learning. In the first to $30$-th iterations, most collected data are negative. The model becomes certain that most data are negative, and the top panels show that even the most likely image has a positive probability of around $0.5$. It indicates that most samples have positive probabilities of less than $0.5$. As Figure \ref{fig: entillust} shows, in this regime, maximizing the acquisition function is equivalent to finding the most likely image to be positive. Seafaring maximizes the positive probability of the data being labeled. After some trials, Seafaring finds out some positive data, and the ratio of negative and positive samples decreases. After the $30$-th iteration, the model's output probability gets back to normal, and Seafaring focuses on retrieving the most uncertain data.

We did not expect this phenomenon. We considered that we needed an explicit mechanism to retrieve positive data in the initial phase. However, it turned out that such a mechanism was unnecessary. This finding let us keep Seafaring simple and easy to implement.

\section{Discussions} \label{sec: discussion}

In the experiments, we used the Flickr database as the pool. As we mentioned earlier, our algorithm is general and can be combined with other search engines and Web services, such as Instagram, DeviantArt, and Google Image Search. The most attractive choice would be general-purpose image search engines, such as Google Image Search and Bing Image Search, because they retrieve images from the entire Web, and the pool becomes even larger. We did not use them because of their cost. For example, Google Search API charges 5 dollars per 1000 queries\footnote{\url{https://developers.google.com/custom-search/v1/overview}}, and Bing Image Search API charges 4 to 7 dollars per 1000 queries\footnote{\url{https://azure.microsoft.com/ja-jp/pricing/details/cognitive-services/search-api/}}. As we use $100$ to $1000$ queries per iteration and $100$ iterations for each run of the experiments, each run would take 40 to 500 dollars for the search API. These API costs are much higher than labeling costs in most applications, and it makes these APIs not attractive in most situations. By contrast, Flickr API offers $3600$ queries per hour for free\footnote{\url{https://www.flickr.com/services/developer/api/}}. Therefore, the entire process of active learning takes $3$ to $30$ hours and is for free, and the bottleneck of the computational cost is training deep models, and the bottleneck of the expense is labeling, which is inevitable in all tasks. Besides, Flickr API can control the licenses of the result images. For example, specifying ``\texttt{license=9, 10}'' restricts the results to public domain images, which resolves license issues. These are why we used Flickr in the experiments and why Flickr API is attractive in most applications. However, although Flickr hosts more than ten billion images, it lacks some types of images, such as illustrations and celebrity portraits. If the task at hand is relevant to such data, other databases, such as Instagram and DeviantArt, should be used. One of the interesting future works is to combine many search engines and services to make Seafaring more robust while keeping the API cost low. An attractive future avenue is a cost-sensitive approach that basically uses ``free'' databases such as Flickr and uses ``costly'' targets such as Google Image Search only when the costly APIs are promising, which alleviates the API cost issue.

We emphasize that Seafaring is applicable to any domain only if an API takes a tag as input and outputs data samples. For example, \texttt{/2/tweets/search/recent} API of Twitter can be used for NLP tasks, such as sentiment analysis and tweet recommendation tasks, where a hashtag is a tag, and a tweet is a data sample.

\begin{figure}[tb]
\centering
\includegraphics[width=0.95\hsize]{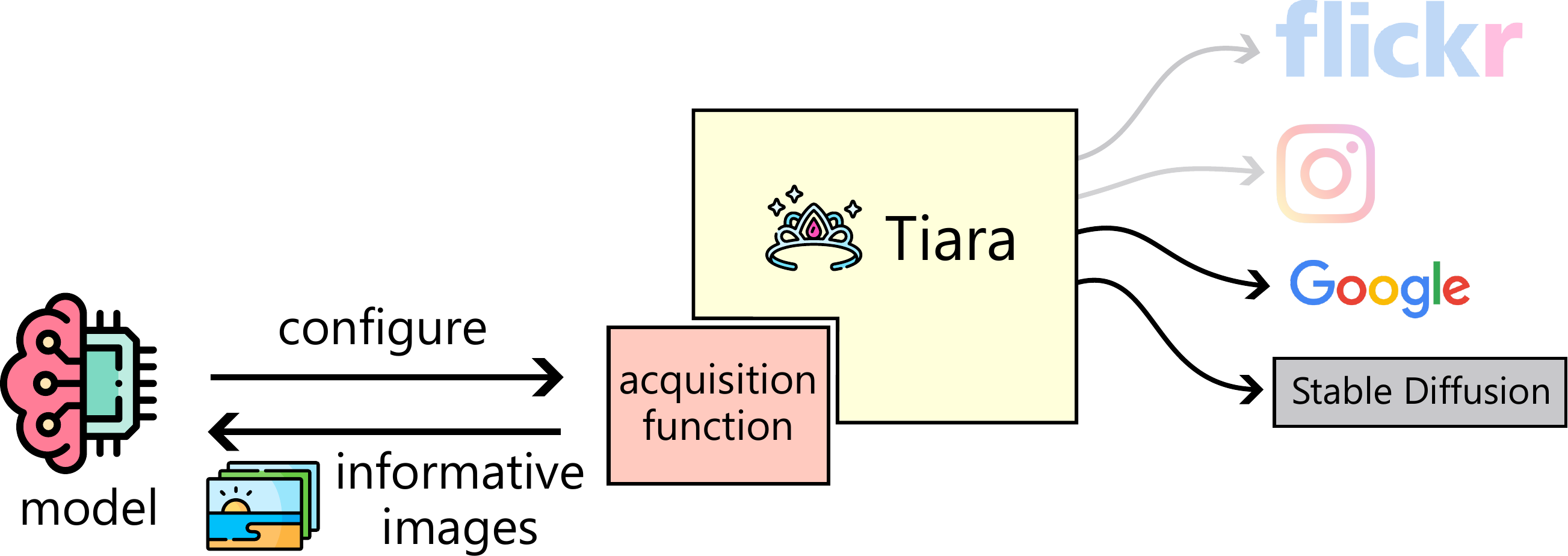}
\caption{\textbf{Future work.} Combining Seafaring with general-purpose search engines such as Google Image Search is interesting because it makes the pool even larger, say trillions of data. However, the API is costly, and they are not economical for many tasks currently. Seafaring can also be combined with generative models such as Stable Diffusion. Integrating Seafaring with such systems is an interesting future direction.}
\label{fig: future}
\vspace{-0.2in}
\end{figure}

Another type of interesting target ``database'' is deep generative models (Figure \ref{fig: future}). Recently, high-quality text-to-image models have been developed and released \cite{rombach2022high,dhariwal2021diffusion}. They generate photo-realistic images and high-quality illustrations based on input text. Tiara and Seafaring can search images from any type of system only if they receive text and provide images, and Seafaring can be combined with such generative models. If we host these models locally, we save communication costs to collect unlabelled images and may be able to boost the performance of Seafaring further.

\section{Related Work}

\subsection{Active Learning}

The aim of active learning is to reduce the burden of labeling data \cite{settles2009active}. Active learning methods try to find out informative unlabelled data so that the total number of required labels is minimized while the performance of the trained model is maximized. There are many problem settings for active learning. In the stream-based tasks \cite{zliobaite2014active,attenberg2011online}, unlabelled data are shown one by one, and the active learning method decides whether we should label each data online. The membership query synthesis problem \cite{angluin1987queries,zhu2017generative} asks active learning methods to generate data from scratch to be labeled. Pool-based active learning is the most popular setting in the literature \cite{tong2001support,gal2017deep,wang2017cost,kirsch2019batchbald,beluch2018power}, and we have focused on it in this paper. In pool-based active learning, a set of unlabelled data is shown, and active learning methods select some of them at once or iteratively. Although Seafaring is classified as a pool-based method, it is interesting to regard Seafaring as the middle of query synthesis and pool-based methods. The pool of Seafaring is extremely large and dense in the input space $\mathcal{X}$, and many new images are added to the pool every day. So the pool contains many images that we do not expect in advance, and searching from the pool shares preferable properties with query synthesis in terms of novelty and diversity. The important difference between Seafaring and query synthesis methods is that all of the images in the pool are indeed real-world data, while query synthesis may generate non-realistic data or corrupted data. If we use generative models as the target database, as we discuss in Section \ref{sec: discussion}, this difference becomes more nuanced.

The criterion on which active learning selects data is important, and there have been many studies on this topic. Typically, the criterion is represented by acquisition function $a\colon \mathcal{X} \to \mathbb{R}$, and active learning methods select the item with the highest acquisition value. The most popular acquisition functions are based on the uncertainty of the model \cite{gal2017deep,wang2017cost} or the amount of disagreement between ensemble models \cite{beluch2018power}. Some methods adaptively design the policy of active learning \cite{fang2017learning,haussmann2019deep}. In either case, the core idea is that uncertain data should be prioritized.

It is important to accelerate an iteration of active learning and reduce the number of iterations to make active learning efficient. A popular approach is batch active learning \cite{hoi2006batch,sener2018active,ash2020deep,shui2020deep,citovsky2021batch}, which allows learners to select multiple instances per iteration and reduces the number of iterations. Batch active learning has been successfully applied to large-scale problems and scales up to a large pool of ten million images and one million labels \cite{citovsky2021batch}. However, they require many labels, and labeling millions of data is not economical or infeasible in many tasks. Besides, this approach does not scale to billion-size pools so far.

The most significant difference between our proposed method and existing active learning lies in the size of the pool of unlabelled data. In typical settings, the pool contains thousands to millions of data, while we consider the pool with billions of data. If the pool does not contain informative data at all, active learning fails. Our huge pool avoids such cases.

\subsection{User-side algorithms}

User-side algorithms, such as user-side recommender systems \cite{sato2022private,sato2022principled} and search systems \cite{bra1994information,sato2022retrieving,sato2022clear}, are designed for users of Web services and software. This is in contrast to typical algorithms, which are designed for official developers of Web services and software. User-side information retrieval algorithms enable us to retrieve data from external databases with respect to user-designed score functions without privileged access to the databases.

Our proposed method uses an off-the-shelf user-side algorithm as a building block. The difference between our proposed method and existing works is its usage. The original motivation for the study of user-side algorithms is to create customized search engines for every user when the official system is not satisfactory. By contrast, we use them for gathering informative data for active learning.

\subsection{Web Mining}

Another relevant realm is Web mining. As Seafaring retrieves data from the Web, it can be seen as a Web mining method. The most relevant technique is focused crawling \cite{chakrabarti1999focused, mccallum2000automating, johnson2003evolving, baezayates2005crawling, guan2008guide,pham2019bootstrapping}, which aims at finding Web pages that meet some criteria by crawling the Web. There are various tasks in focused crawling depending on the criteria. For example, the target can be pages of specific topics \cite{chakrabarti1999focused, mccallum2000automating}, popular pages \cite{baezayates2005crawling}, structured data \cite{meusel2014focused}, and hidden pages \cite{barbosa2007adaptive}. To the best of our knowledge, there are no focused crawling methods to retrieve data for active learning.

We adopted an approach using a user-side algorithm instead of focused crawling. The advantage of user-side algorithms is that it works online and retrieves data in a few seconds to a few minutes, whereas focused crawling is typically realized by a resident program and takes a few hours to a few weeks, in which case the crawling procedure is a bottleneck of the time consumption. As active learning involves hundreds of iterations of data acquisition, it would be too expensive if each iteration took hours to weeks. With that being said, when the time constraint is not severe, the focused crawling approach is an appealing candidate because it does not rely on off-the-shelf APIs but can be applied to any Web pages. 

Another relevant realm is dataset mining from the Web \cite{zhang2021dsdd,castelo2021auctus}. These methods retrieve sets of compiled data, such as zip files and JSON files, while Seafaring retrieves unlabeled data from databases that are not designed for training models. The dataset mining methods and engines are easier to use and attractive if the target task is clear, while Seafaring can be used even when the task is not solid.

\section{Conclusion}

In this paper, we have proposed Seafaring, an active learning method that uses a huge pool of unlabelled data. Compared with existing active learning research, Seafaring uses a several orders of magnitude larger pool. As the pool is extremely large, it is more likely that relevant images exist in the pool. Besides, the pool is general and contains diverse images, we use the same pool for many tasks, and we do not need to design the unlabelled pool for each task. So Seafaring reduces the cost of active learning further. The pool is, in addition, always up-to-date without any cost of maintenance because new images are added to the pool by the users of Flickr. In the experiments, we show that Seafaring is more effective than the counterpart of Seafaring with a smaller database and show the importance of the size of the pool. We also show that Seafaring automatically balances the ratio of positive and negative data without any explicit measures, while random sampling tends to have negative samples.

\begin{acks}
This work was supported by JSPS KAKENHI GrantNumber 21J22490. 
\end{acks}


\bibliographystyle{plainnat}
\bibliography{sample-base}

\clearpage

\appendix








\end{document}